\documentclass{article} 
\usepackage{iclr2025_conference,times}

\usepackage{amsmath,amsfonts,bm}

\def\eqref#1{equation~\ref{#1}}

\def\1{\bm{1}}

\DeclareMathAlphabet{\mathsfit}{\encodingdefault}{\sfdefault}{m}{sl}
\SetMathAlphabet{\mathsfit}{bold}{\encodingdefault}{\sfdefault}{bx}{n}

\usepackage{hyperref}
\usepackage{url}

\usepackage{float}
\usepackage{graphicx}
\usepackage{subfigure}
\usepackage{wrapfig}
\usepackage{amsmath}
\usepackage{amsfonts}
\usepackage{amsthm}

\newtheorem{theorem}{Theorem}

\usepackage{bm}
\usepackage{tabularx}
\usepackage{booktabs}
\usepackage{setspace}
\usepackage{xcolor,colortbl}

\definecolor{lightblue}{rgb}{0.92, 0.95, 1}
\definecolor{lightred}{rgb}{1, 0.9, 0.9}
\definecolor{deepblue}{rgb}{0, 0.4470, 0.7410}
\definecolor{deepyellow}{rgb}{0.9290, 0.6940, 0.1250}
\definecolor{deepgreen}{rgb}{0,0.5,0}
\usepackage{enumitem}
\setlist[itemize]{left=1em}

\title{KAE: Kolmogorov-Arnold Auto-Encoder for Representation Learning}

\author{Fangchen Yu$^1$, Ruilizhen Hu$^1$, Yidong Lin$^1$, Yuqi Ma$^1$, Zhenghao Huang$^1$, Wenye Li$^{1,2}$\thanks{Corresponding author.} \\
$^1$ The Chinese University of Hong Kong, Shenzhen \\
$^2$ Shenzhen Research Institute of Big Data
}

\iclrfinalcopy 
\begin{document}

\maketitle

\begin{abstract}
The Kolmogorov-Arnold Network (KAN) has recently gained attention as an alternative to traditional multi-layer perceptrons (MLPs), offering improved accuracy and interpretability by employing learnable activation functions on edges. In this paper, we introduce the Kolmogorov-Arnold Auto-Encoder (KAE), which integrates KAN with autoencoders (AEs) to enhance representation learning for retrieval, classification, and denoising tasks. Leveraging the flexible polynomial functions in KAN layers, KAE captures complex data patterns and non-linear relationships. Experiments on benchmark datasets demonstrate that KAE improves latent representation quality, reduces reconstruction errors, and achieves superior performance in downstream tasks such as retrieval, classification, and denoising, compared to standard autoencoders and other KAN variants. These results suggest KAE's potential as a useful tool for representation learning. Our code is available at \url{https://github.com/SciYu/KAE/}.

\end{abstract}


\section{Introduction}

Autoencoders \citep{hinton2006reducing,berahmand2024autoencoders} are a fundamental component of modern deep learning, serving as powerful tools for unsupervised representation learning. By compressing input data into a lower-dimensional latent space and subsequently reconstructing it, autoencoders facilitate a wide range of applications, including dimensionality reduction \citep{wang2016auto,lin2020deep}, image classification \citep{zhou2019learning,zhou2023self}, and data denoising \citep{gondara2016medical,cui2024high}. Their ability to learn meaningful representations from unlabelled data has positioned them as essential in various AI domains, from computer vision \citep{mishra2018generative,takeishi2021physics} to natural language processing \citep{shen2020educating,kim2021conditional}, significantly enhancing the performance of downstream tasks.

Traditional autoencoders typically leverage multi-layer perceptrons (MLPs), characterized by fully connected layers. In a conventional autoencoder, each layer computes its output as $y=\sigma(Wx+b)$, where $\sigma$ denotes a fixed activation function, $W$ is the learnable weight matrix, $x$ represents the input, and $b$ is the bias vector. The optimization of $W$ is traditionally performed through black-box AI systems, limiting the adaptability of the activation function to the data’s underlying structure.

Recent advancements in alternative architectures, such as Kolmogorov-Arnold Networks (KANs) \citep{liu2024kan,liu2024kan2}, offer a compelling pathway for enhancing autoencoder performance. Unlike MLPs, KANs redefine the output as $y=f(x)$, where $f$ is a learnable activation function. This shift enables the network to adaptively learn more complex representations, moving beyond the limitations of fixed functions and linear weights used in MLPs.

In this paper, we propose the Kolmogorov-Arnold Auto-Encoder (KAE), which integrates the strengths of KANs into the autoencoder framework to create a more robust model for representation learning. However, merely substituting MLP layers with KAN layers utilizing B-spline functions may result in only marginal gains or even a decline in performance. The efficacy of KANs heavily depends on the specific type of function employed within the KAN layer. To address this challenge, we introduce a well-defined KAE architecture that utilizes learnable polynomial functions, demonstrating promising improvements in representation learning.

Our contributions are as follows:
\begin{itemize}
    \item We introduce the Kolmogorov-Arnold Representation Theorem into the design of autoencoders, providing a theoretical basis for improving autoencoder performance.
    
    \item We investigate the role of activation functions in the KAN layers, identifying polynomial functions as a suitable choice for enhancing autoencoder performance.
    
    \item We demonstrate the superiority of the Kolmogorov-Arnold Auto-Encoder (KAE) through extensive experimental validation, showing improvements in both reconstruction quality and downstream application performance.
\end{itemize}

The remainder of the paper is organized as follows. Section \ref{sec:related} reviews related work. Section \ref{sec:model} details the proposed KAE model, and Section \ref{sec:evaluation} provides a thorough empirical evaluation. Finally, Section \ref{sec:conclusion} concludes the paper.


\section{Related Work}
\label{sec:related}


\subsection{Autoencoders (AEs)}
\label{sec:related:autoencoder}

\begin{figure}[H]
    \centering
    \includegraphics[width=.98\textwidth]{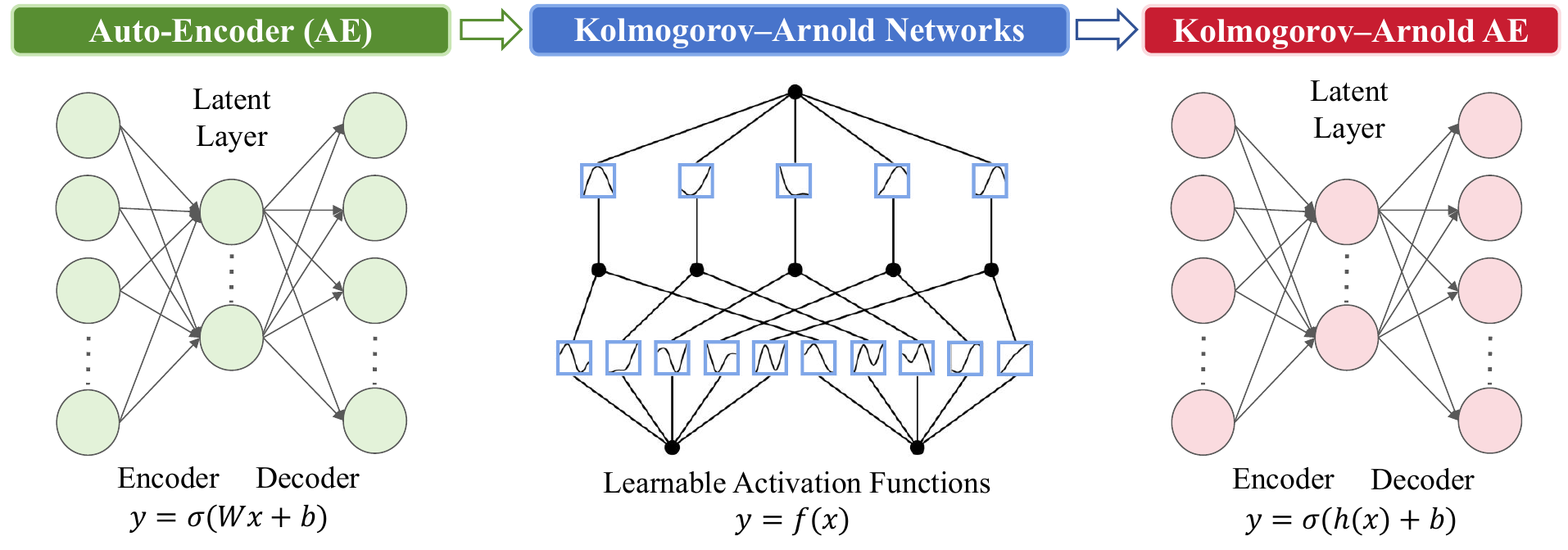}
    \caption{\textbf{Model Comparison of AE, KAN, and KAE.}}
    \label{fig:models}
\end{figure}

Autoencoders \citep{hinton2006reducing}, as a framework of unsupervised learning, aim to learn a compressed representation of input data while minimizing reconstruction error. Traditionally, autoencoders rely on MLPs to achieve this task, where each layer applies a fixed non-linearity. However, this may limit the ability of the network to capture complex structures in the data. 

Recent works have explored various autoencoder architectures, including Variational Auto-Encoders (VAEs) \citep{kingma2013vae,kingma2019introduction} and Denoising Auto-Encoders (DAEs) \citep{savinov2022dae,wu2023dae}, which introduce stochasticity and robustness to noise, respectively. Nevertheless, most of these architectures rely on fixed activation functions, which may limit their flexibility in representing complex functions.

As shown in Fig. \ref{fig:models}, an autoencoder consists of two main parts:
\begin{itemize}
    \item \textbf{Encoder}: This part compresses the input into a latent-space representation.
    \item \textbf{Decoder}: This part reconstructs the input data from the latent-space representation.
\end{itemize}

A common application of autoencoders is representation learning, where latent representations are learned and can be used for tasks such as image classification \citep{zhou2023self}, text classification \citep{xu2019semisupervised}, cross-modal learning \citep{dong2023}, and other applications, particularly beneficial for high-dimensional data analysis. Another key feature of autoencoders is their ability to reduce noise in data \citep{bajaj2020autoencoders}. By inputting noisy data, the pre-trained autoencoder can remove the noise, producing clean outputs, which makes it a powerful tool for data denoising.


\clearpage
\subsection{Kolmogorov-Arnold Networks (KANs)}

Kolmogorov-Arnold Networks (KANs) \citep{liu2024kan,liu2024kan2} provide a flexible way to model non-linear relationships by using learnable activation functions rather than fixed ones. Inspired by the Kolmogorov-Arnold representation theorem \citep{kolmogorov1961representation,braun2009constructive}, KANs approximate any continuous multivariate function using a sum of univariate functions, providing a theoretically grounded approach to function approximation. In practice, KANs allow each layer to learn its own activation function, making them adaptable to non-linear data. KANs have been shown to outperform MLPs in various applications \citep{xu2024fourierkan,bozorgasl2024wavkan}.

\subsubsection{Kolmogorov-Arnold Representation Theorem}
\label{sec:related:kolmogorov}

The Kolmogorov-Arnold representation theorem \citep{kolmogorov1961representation,braun2009constructive} is a fundamental result in mathematics, particularly in functional analysis and multivariate approximation theory. It provides key insights into representing multivariate continuous functions. Theorem~\ref{thm:KA} asserts that any continuous function of $d$ variables can be represented as a finite sum of continuous univariate functions along with an additional continuous function.
\begin{theorem}[\textbf{Kolmogorov-Arnold Representation Theorem}] \label{thm:KA}
    For any smooth function $f: [0,1]^d \rightarrow \mathbb{R}$, there exist continuous functions $\phi_{k,j}: [0,1] \rightarrow \mathbb{R}$ and $\Phi_k: \mathbb{R} \rightarrow \mathbb{R}$ such that:    
    \begin{equation} \label{eq:KA_theorem}
        f(x_1, x_2, \ldots, x_d) = \sum_{k=1}^{2d+1} \Phi_k \left( \sum_{j=1}^{d} \phi_{k,j}(x_j) \right).
    \end{equation}
\end{theorem}
This remarkable theorem involves the use of two primary types of functions: \textbf{inner functions} $\{\phi_{k,j}\}$ and \textbf{outer functions} $\{\Phi_k\}$, which motivates the design of the KAN network. 

\subsubsection{KAN Network and Its Applications}

Inspired by Theorem~\ref{thm:KA}, \cite{liu2024kan,liu2024kan2} proposed a novel KAN layer with $d_\text{in}$-dimensional inputs and $d_\text{out}$-dimensional outputs, defined as a matrix of 1D functions:
\[ \bm{\Phi} := \{\phi_{k,j}\}, \quad j=1,2,\cdots,d_{\text{in}}, \quad k=1,2,\cdots,d_{\text{out}}, \]
where each function $\phi_{k,j}$ is a learnable univariate function. A general KAN network is formed by stacking $L$ KAN layers. Given an input vector $x \in \mathbb{R}^d$, the output of KAN is
\begin{equation}
    \text{KAN}(x) := (\Phi_{L-1} \circ \Phi_{L-2} \circ \cdots \circ \Phi_1 \circ \Phi_0) \circ x,
\end{equation}
where the Kolmogorov-Arnold representation in Eq.~(\ref{eq:KA_theorem}) can be viewed as a composition of two KAN layers: $\Phi_0$ contains $(2d+1) \cdot d$ functions, and $\Phi_1$ contains $1 \cdot (2d+1)$ functions.

Recently, KAN networks have been applied to various domains, including scientific discovery \citep{liu2024kan,liu2024kan2}, image segmentation \citep{li2024ukan}, image classification \citep{cheon2024kolmogorov}, text classification \citep{imran2024leveraging}, collaborative filtering \citep{xu2024fourierkan}, and others. KAN-based architectures, such as the Kolmogorov-Arnold Transformer \citep{yang2024kat} and U-KAN \citep{li2024ukan}, have also gained significant attention.


\section{Kolmogorov–Arnold Auto-Encoder}
\label{sec:model}

\subsection{KANs in Autoencoders}
KANs have demonstrated potential for enhancing neural network architectures, but their application to autoencoders poses challenges. Specifically, the complexity introduced by learnable activation functions can result in overfitting or suboptimal performance if not carefully managed. To address this, we propose a novel autoencoder architecture that integrates KANs with polynomial-based activation functions, which we found to be more stable and effective for this task.

A concurrent work \citep{moradi2024kolmogorov} also introduces a KAN-based autoencoder. However, their approach employs B-spline functions in the KAN layer and adopts a KAN-ReLU-Dense structure in the encoder. In contrast, our architecture uses polynomial functions in the KAN layer, significantly reducing the number of hyperparameters and eliminating the need for dense layers. This design offers a more efficient and streamlined alternative.

\subsection{Overall Architecture}
\label{sec:model:encoder}

The key idea behind the Kolmogorov-Arnold Auto-Encoder (KAE) is to replace the MLP layers (fixed activation functions) in traditional autoencoders with KAN layers (learnable functions), as inspired by the Kolmogorov-Arnold representation theorem. 

In a traditional autoencoder, the MLP consists of fully connected layers that apply a fixed activation function, such as ReLU or sigmoid, after a linear transformation. Given an input vector $x \in \mathbb{R}^d$, the encoder compresses the data into a lower-dimensional latent representation $z \in \mathbb{R}^k$, and the decoder reconstructs $z$ back into the original space. This transformation is typically expressed as:
\[
z = \sigma(Wx + b),
\]
where $W$ is the weight matrix, $b$ is the bias, and $\sigma$ is a fixed activation function.

In KAE, we replace the MLP layers with KAN layers, which use learnable activation functions as defined by the Kolmogorov-Arnold representation theorem. Instead of a fixed $\sigma$, each KAN layer dynamically learns its own activation function, enabling the model to capture complex, non-linear patterns. The encoder applies a series of KAN layers to map the input to the latent space:
\[
z = f(x) := (\Phi_{L-1} \circ \Phi_{L-2} \circ \cdots \circ \Phi_1 \circ \Phi_0) \circ x,
\]
where each $\Phi_i$ represents a KAN layer composed of learnable univariate functions.

Similarly, the decoder ($y=g(z)$) uses KAN layers in reverse to reconstruct the data, providing greater flexibility and accuracy compared to standard MLP-based decoders. This architecture enables KAE to capture complex, non-linear relationships, enhancing the quality of the learned representations. As with standard autoencoders, we use the mean squared error (MSE) between the original and reconstructed data as the training loss.


\subsection{Polynomial Activation Function}

The autoencoder's primary task is to ensure that the encoding and decoding functions are inverses of each other, i.e., $x \approx g(f(x))$ for all $x \in \mathbb{R}^d$. In a traditional autoencoder, the layers are fully connected, and this inversion property is straightforward to achieve. However, when using KAN layers, the choice of functions $f$ and $g$ becomes more nuanced. Na\"{i}vely replacing the MLP layer with a KAN layer does not produce satisfactory results, as shown in our evaluation in Section \ref{sec:evaluation}.

To ensure consistency between the encoder and decoder, we carefully design the $f(x)$ and $g(z)$ to be invertible, using polynomial approximations that are both smooth and differentiable. For a KAE layer with $d_\text{in}$-dimensional inputs and $d_\text{out}$-dimensional outputs, the output of $x \in \mathbb{R}^{d_\text{in}}$ is defined as:
\begin{equation}
    \text{KAE}(x) := \sigma(h(x) + b) = \sigma((c_0 1_{d_\text{in}} + c_1x + c_2x^2 + \dots + c_px^p) + b),
\end{equation}
where $1_{d_\text{in}}$ is an all-ones vector and $p$ is the order of the polynomials, treated as a hyperparameter in KAE. To maintain consistency with the MLP-based AE, we use the structure $\sigma(\cdot + b)$, replacing the $Wx$ term with $h(x)$, resulting in $\sigma(h(x) + b)$ with the following advantages.

\begin{itemize}
    \item Compared to the linear transformation $Wx$ in MLPs, polynomial functions $h(x)$ introduce higher-order non-linear terms such as $x^p$, enabling the model to capture more complex, non-linear relationships in the data.
    
    \item The polynomial function includes a constant matrix $c_0 \in \mathbb{R}^{d_\text{out} \times d_\text{in}}$, providing more flexibility than the traditional bias term $b \in \mathbb{R}^{d_\text{out}}$, allowing the model to better adapt to shifts and variations in the data.
\end{itemize}

Notably, in the standard KAN \citep{liu2024kan}, the learnable function $h(x)$ is set as a B-spline function, while in FourierKAN \citep{xu2024fourierkan} it is set as a Fourier function, and in WavKAN \citep{bozorgasl2024wavkan} it is set as a Wavelet function. Comparatively, our empirical validation shows that quadratic and cubic polynomial functions (with $p=2,3$) offer a more effective balance between flexibility and stability, improving the model's ability to reconstruct data while maintaining the inversion property between the encoder and decoder.


\section{Evaluation}
\label{sec:evaluation}

\subsection{Experimental Setup}
\label{sec:evaluation:settings}

To assess the effectiveness of the proposed Kolmogorov-Arnold Auto-Encoder (\textbf{KAE}), we conducted a series of experiments to address the following key objectives:
\begin{itemize}
    \item \textbf{Reconstruction in representation learning}: Evaluating the model's ability to effectively compress and decompress data by comparing reconstructed outputs to the original inputs.

    \item \textbf{Quality of latent representations}: Assessing the utility of learned representations in downstream tasks such as similarity search, image classification, and image denoising.

    \item \textbf{Model capacity}: Analyzing the model's performance relative to the number of learnable parameters, providing insight into its capacity and efficiency. 
\end{itemize}

We compared the proposed KAE\footnote{The implementation of KAE is provided at \url{https://github.com/SciYu/KAE}, which is built upon the TaylorKAN codebase (\url{https://github.com/Muyuzhierchengse/TaylorKAN}). In our experiments, the polynomial order \(p\) of the KAE is set to 1, 2, and 3, with the Sigmoid function used as $\sigma$.} against the following baseline models: 

\begin{itemize}
    \item \textbf{Standard Auto-Encoder (AE)} \citep{hinton2006reducing}: Utilizes the Sigmoid activation function.  

    \item \textbf{Kolmogorov-Arnold Network (KAN)} \citep{liu2024kan,liu2024kan2}: Employs B-spline activation functions. The default parameters are consistent with the Efficient-KAN implementation\footnote{\url{https://github.com/Blealtan/efficient-kan}}, such as \texttt{grid\_size = 5}, \texttt{spline\_order = 3}, and \texttt{grid\_range = [-1,1]}.

    \item \textbf{FourierKAN} \citep{xu2024fourierkan}: Uses the default setting of the basic implementation of FourierKAN\footnote{\url{https://github.com/GistNoesis/FourierKAN}} with \texttt{grid\_size = 5}, adds bias, and does not apply smooth initialization.

    \item \textbf{WavKAN} \citep{bozorgasl2024wavkan}: Implements the default configuration as the official code\footnote{\url{https://github.com/zavareh1/Wav-KAN}}, including Mexican hat wavelets as activation functions.
\end{itemize}

We employed several well-known image datasets for our evaluations, as summarized in Table \ref{tab:dataset}.
\vspace{-2mm}

\begin{table}[H]
\centering
\caption{\textbf{Statistics of the Image Datasets Used in Our Work.}}
\label{tab:dataset}
\setlength{\tabcolsep}{1pt}
\small
\begin{spacing}{1.1}
\begin{tabular}{lcccccccc}
\toprule
Dataset & Image Type & Image Size & \#Classes & \#Training & \#Test \\ \hline
MNIST \citep{lecun1998mnist} & Grayscale handwritten digits & 28$\times$28 & 10 & 60,000 & 10,000 \\ 
FashionMNIST \citep{xiao2017fashionmnist} & Grayscale images of clothing & 28$\times$28 & 10 & 60,000 & 10,000 \\ 
CIFAR10 \citep{2009cifar} & RGB natural images & 32$\times$32 & 10 & 50,000 & 10,000 \\ 
CIFAR100 \citep{2009cifar} & RGB natural images & 32$\times$32 & 100 & 50,000 & 10,000 \\ 
\bottomrule
\end{tabular}
\end{spacing}
\end{table}

\vspace{-2mm}

Each experiment was repeated ten times with random seeds from 2,024 to 2,033, and the average results with standard deviation were reported. Models were trained using the Adam optimizer \citep{kingma2014adam}, exploring four configurations of learning rate (1e-4 or 1e-5) and weight decay (1e-4 or 1e-5), with a batch size of 256 for 10 epochs. The best-performing configuration was reported.

All experiments were conducted using Python (version 3.10) and PyTorch 2.4 as the deep learning framework. Computations were performed on a ThinkStation equipped with an Intel i7-12700 CPU (2.1 GHz), 32GB of RAM, and an NVIDIA TITAN V GPU with 12GB of GPU memory. 


\subsection{Reconstruction Quality}
\label{sec:evaluation:compression}

Autoencoders perform the representation learning by compressing input data into a latent space (encoding) and then reconstructing it back (decoding). To assess how well different models perform this task, we compared their reconstruction error using the mean squared error (MSE) between the original input and the reconstructed output on the test set.

For all models, we employed a shallow architecture consisting of three layers: $d_{\text{input}}$-$d_{\text{latent}}$-$d_{\text{output}}$, where $d_{\text{latent}}$ represents the dimension of the compressed latent space. After training each model, a previously unseen test input $x$ was passed through the network to obtain the latent representation $y$ and the reconstructed output $z$. The reconstruction error was then calculated as the MSE, defined by $\left\| x-z \right\|^2$, and averaged across all test data batches.

Table~\ref{tab:reconstruction} demonstrates the superiority of KAE compared to AE and KAN variants in terms of reconstruction error. The results highlight several key observations:
\begin{itemize}
    \item \textbf{KAE vs AE}: Both KAN and KAE consistently deliver significantly lower reconstruction errors than AE in all settings, with the error reduction of KAE highlighted in the last row of Table~\ref{tab:reconstruction}.

    \item \textbf{KAE vs KAN}: The performance of KAN variants is highly dependent on the choice of activation function. The polynomial function used in our KAE model outperforms the B-spline function in KAN, the Fourier function in FourierKAN, and the wavelet function in WavKAN for this reconstruction task. Additionally, higher-order polynomial functions (i.e., $p=3$) in KAE lead to better reconstruction performance.
\end{itemize}
\vspace{-2mm}

\begin{table}[H]
\centering
\caption{\textbf{Reconstruction Error Comparison Across Datasets for Different Latent Dimensions.} The best results are in \textbf{bold} and the second best are \underline{underlined}. The last row shows the improvement of KAE models over standard autoencoders (AE).}
\label{tab:reconstruction}
\setlength{\tabcolsep}{1.7pt}
\small
\begin{spacing}{1.1}
\begin{tabular}{lcccccccc}
\toprule
Dataset & \multicolumn{2}{c}{MNIST} & \multicolumn{2}{c}{FashionMNIST} & \multicolumn{2}{c}{CIFAR10} & \multicolumn{2}{c}{CIFAR100} \\ 
\cmidrule(rr){2-3} \cmidrule(rr){4-5} \cmidrule(rr){6-7} \cmidrule(rr){8-9}
Dimension & 16 & 32 & 16 & 32 & 16 & 32 & 16 & 32 \\ \hline
AE & 0.056$_{\pm \text{0.002}}$ & 0.043$_{\pm \text{0.001}}$ & 0.045$_{\pm \text{0.002}}$ & 0.034$_{\pm \text{0.001}}$ & 0.034$_{\pm \text{0.001}}$ & 0.029$_{\pm \text{0.001}}$ & 0.037$_{\pm \text{0.001}}$ & 0.030$_{\pm \text{0.001}}$ \\
KAN & 0.047$_{\pm \text{0.002}}$ & 0.036$_{\pm \text{0.001}}$ & 0.032$_{\pm \text{0.003}}$ & 0.024$_{\pm \text{0.000}}$ & 0.025$_{\pm \text{0.001}}$ & 0.019$_{\pm \text{0.000}}$ & 0.025$_{\pm \text{0.001}}$ & 0.018$_{\pm \text{0.001}}$ \\
FourierKAN & 0.042$_{\pm \text{0.003}}$ & 0.031$_{\pm \text{0.003}}$ & 0.031$_{\pm \text{0.001}}$ & 0.024$_{\pm \text{0.001}}$ & 0.031$_{\pm \text{0.002}}$ & 0.023$_{\pm \text{0.001}}$ & 0.029$_{\pm \text{0.001}}$ & 0.022$_{\pm \text{0.001}}$ \\
WavKAN & 0.175$_{\pm \text{0.002}}$ & 0.161$_{\pm \text{0.002}}$ & 0.099$_{\pm \text{0.001}}$ & 0.089$_{\pm \text{0.000}}$ & 0.035$_{\pm \text{0.001}}$ & 0.025$_{\pm \text{0.000}}$ & 0.036$_{\pm \text{0.001}}$ & 0.026$_{\pm \text{0.000}}$ \\
\rowcolor{lightblue}
KAE (p=1) & 0.050$_{\pm \text{0.004}}$ & 0.041$_{\pm \text{0.000}}$ & 0.029$_{\pm \text{0.001}}$ & 0.025$_{\pm \text{0.001}}$ & 0.021$_{\pm \text{0.000}}$ & 0.020$_{\pm \text{0.000}}$ & 0.021$_{\pm \text{0.001}}$ & 0.019$_{\pm \text{0.000}}$ \\
\rowcolor{lightblue}
KAE (p=2) & \underline{0.026}$_{\pm \text{0.002}}$ & \underline{0.017}$_{\pm \text{0.001}}$ & \underline{0.020}$_{\pm \text{0.000}}$ & \underline{0.016}$_{\pm \text{0.001}}$ & \underline{0.017}$_{\pm \text{0.001}}$ & \underline{0.013}$_{\pm \text{0.000}}$ & \underline{0.017}$_{\pm \text{0.001}}$ & \underline{0.013}$_{\pm \text{0.000}}$ \\
\rowcolor{lightblue}
KAE (p=3) & \textbf{0.024}$_{\pm \text{0.001}}$ & \textbf{0.015}$_{\pm \text{0.001}}$ & \textbf{0.018}$_{\pm \text{0.001}}$ & \textbf{0.015}$_{\pm \text{0.000}}$ & \textbf{0.016}$_{\pm \text{0.001}}$ & \textbf{0.012}$_{\pm \text{0.000}}$ & \textbf{0.016}$_{\pm \text{0.001}}$ & \textbf{0.012}$_{\pm \text{0.000}}$ \\
Improve & \multicolumn{1}{l}{\color{deepgreen}0.032 $\downarrow$} & \multicolumn{1}{l}{\color{deepgreen}0.028 $\downarrow$} & \multicolumn{1}{l}{\color{deepgreen}0.027 $\downarrow$} & \multicolumn{1}{l}{\color{deepgreen}0.019 $\downarrow$} & \multicolumn{1}{l}{\color{deepgreen}0.018 $\downarrow$} & \multicolumn{1}{l}{\color{deepgreen}0.017 $\downarrow$} & \multicolumn{1}{l}{\color{deepgreen}0.021 $\downarrow$} & \multicolumn{1}{l}{\color{deepgreen}0.018 $\downarrow$} \\
\bottomrule
\end{tabular}
\end{spacing}
\end{table}


\subsection{Applications}
\label{sec:evaluation:applications}

To evaluate the quality of the compressed data, we applied the latent representations to several downstream tasks, including similarity search, image classification, and image denoising, demonstrating the high-quality representations learned by the proposed KAE models.

\subsubsection{Similarity Search}

A well-designed autoencoder, as a tool for dimensionality reduction, should preserve the distance relationships between samples from the input space to the latent space, making similarity search a suitable application for assessing this property.

For each experiment, we randomly selected a subset of 1,000 test samples as the test set. For each query, we computed the $k$ nearest neighbors in the input space as the ground truth and compared it to the $N$ nearest neighbors retrieved in the latent space. We calculated the recall, defined as the ratio of true top-$k$ results within the top-$N$ retrieved candidates, and reported it as Recall $k$@$N$. In our experiments, we set $k = 10$ and tested Recall 10@$N$ (referred to as \textbf{Recall@$\bm{N}$}) for all models. Each experiment was repeated 10 times with different random seeds, and we reported the averaged Recall with standard deviations.

Table~\ref{tab:search} presents the Recall@10 results for various models across latent dimensions (16 and 32). Increasing the latent dimension from 16 to 32 significantly improved retrieval recall for all models, enhancing their distance-preserving properties and expressive power. Notably, the proposed KAE model consistently outperformed both the standard AE and KAN variants across all datasets and dimensions. Specifically, KAE ($p=2$) achieved improvements of 0.242 and 0.206 in Recall@10 over AE for MNIST with latent dimensions of 16 and 32, respectively, with similar gains observed for FashionMNIST, CIFAR10, and CIFAR100 datasets.

\begin{table}[H]
\centering
\caption{\textbf{Retrieval Recall@10 Comparison Across Datasets for Different Latent Dimensions.} The best results are in \textbf{bold} and the second best are \underline{underlined}. The last row shows the improvement of KAE models over standard autoencoders (AE).}
\label{tab:search}
\setlength{\tabcolsep}{1.7pt}
\small
\begin{spacing}{1.1}
\begin{tabular}{lcccccccc}
\toprule
Dataset & \multicolumn{2}{c}{MNIST} & \multicolumn{2}{c}{FashionMNIST} & \multicolumn{2}{c}{CIFAR10} & \multicolumn{2}{c}{CIFAR100} \\ 
\cmidrule(rr){2-3} \cmidrule(rr){4-5} \cmidrule(rr){6-7} \cmidrule(rr){8-9}
Dimension & 16 & 32 & 16 & 32 & 16 & 32 & 16 & 32 \\ \hline
AE & 0.354$_{\pm \text{0.017}}$ & 0.483$_{\pm \text{0.008}}$ & 0.401$_{\pm \text{0.016}}$ & 0.526$_{\pm \text{0.009}}$ & 0.368$_{\pm \text{0.026}}$ & 0.472$_{\pm \text{0.012}}$ & 0.380$_{\pm \text{0.015}}$ & 0.477$_{\pm \text{0.009}}$ \\
KAN & 0.457$_{\pm \text{0.015}}$ & 0.552$_{\pm \text{0.016}}$ & 0.495$_{\pm \text{0.023}}$ & 0.578$_{\pm \text{0.004}}$ & 0.377$_{\pm \text{0.015}}$ & 0.496$_{\pm \text{0.007}}$ & 0.391$_{\pm \text{0.009}}$ & 0.512$_{\pm \text{0.007}}$ \\
FourierKAN & 0.498$_{\pm \text{0.053}}$ & 0.638$_{\pm \text{0.034}}$ & 0.518$_{\pm \text{0.022}}$ & \underline{0.615}$_{\pm \text{0.018}}$ & 0.258$_{\pm \text{0.041}}$ & 0.406$_{\pm \text{0.028}}$ & 0.316$_{\pm \text{0.026}}$ & 0.435$_{\pm \text{0.019}}$ \\
WavKAN & 0.259$_{\pm \text{0.037}}$ & 0.447$_{\pm \text{0.018}}$ & 0.387$_{\pm \text{0.015}}$ & 0.488$_{\pm \text{0.006}}$ & 0.258$_{\pm \text{0.016}}$ & 0.428$_{\pm \text{0.012}}$ & 0.259$_{\pm \text{0.013}}$ & 0.430$_{\pm \text{0.011}}$ \\
\rowcolor{lightblue}
KAE (p=1) & 0.404$_{\pm \text{0.028}}$ & 0.488$_{\pm \text{0.007}}$ & 0.544$_{\pm \text{0.008}}$ & 0.581$_{\pm \text{0.010}}$ & 0.440$_{\pm \text{0.011}}$ & 0.489$_{\pm \text{0.008}}$ & 0.453$_{\pm \text{0.008}}$ & 0.504$_{\pm \text{0.005}}$ \\
\rowcolor{lightblue}
KAE (p=2) & \textbf{0.596}$_{\pm \text{0.019}}$ & \textbf{0.689}$_{\pm \text{0.011}}$ & \textbf{0.607}$_{\pm \text{0.010}}$ & \textbf{0.661}$_{\pm \text{0.007}}$ & \textbf{0.525}$_{\pm \text{0.021}}$ & \textbf{0.631}$_{\pm \text{0.011}}$ & \textbf{0.544}$_{\pm \text{0.013}}$ & \textbf{0.639}$_{\pm \text{0.007}}$ \\
\rowcolor{lightblue}
KAE (p=3) & \underline{0.554}$_{\pm \text{0.013}}$ & \underline{0.659}$_{\pm \text{0.013}}$ & \underline{0.521}$_{\pm \text{0.006}}$ & 0.582$_{\pm \text{0.006}}$ & \underline{0.488}$_{\pm \text{0.012}}$ & \underline{0.597}$_{\pm \text{0.010}}$ & \underline{0.493}$_{\pm \text{0.013}}$ & \underline{0.586}$_{\pm \text{0.012}}$ \\
Improve & \multicolumn{1}{l}{\color{deepgreen}0.242 $\uparrow$} & \multicolumn{1}{l}{\color{deepgreen}0.206 $\uparrow$} & \multicolumn{1}{l}{\color{deepgreen}0.206 $\uparrow$} & \multicolumn{1}{l}{\color{deepgreen}0.135 $\uparrow$} & \multicolumn{1}{l}{\color{deepgreen}0.157 $\uparrow$} & \multicolumn{1}{l}{\color{deepgreen}0.159 $\uparrow$} & \multicolumn{1}{l}{\color{deepgreen}0.164 $\uparrow$} & \multicolumn{1}{l}{\color{deepgreen}0.162 $\uparrow$} \\
\bottomrule
\end{tabular}
\end{spacing}
\end{table}

\begin{figure}[H]
    \centering
    \subfigure[Latent dimension $d_{\text{latent}}=16$]{
    \includegraphics[width=0.99\linewidth]{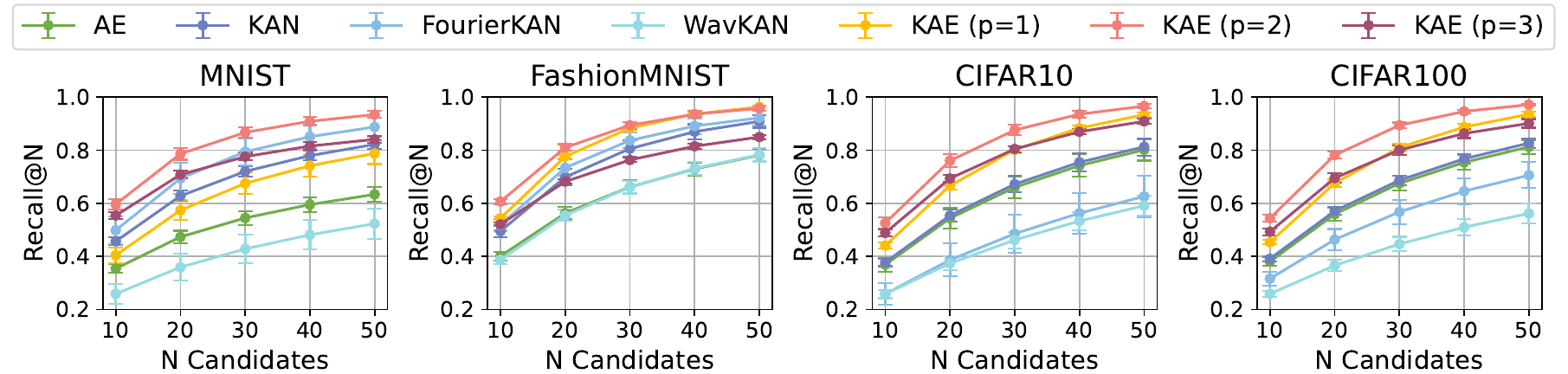}}
    \vspace{-2mm}
    
    \subfigure[Latent dimension $d_{\text{latent}}=32$]{
    \includegraphics[width=0.99\linewidth]{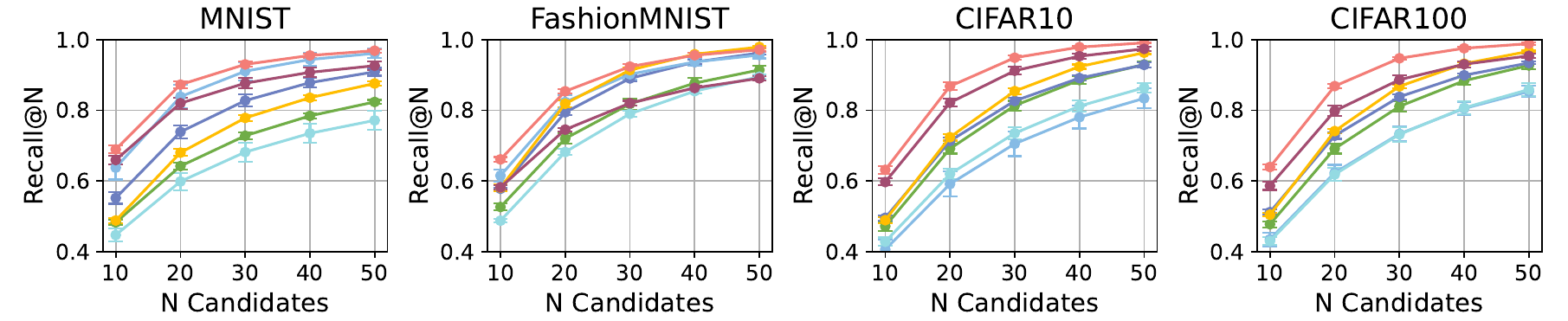}}
    \vspace{-3mm}
    
    \caption{\textbf{Recall@$\bm{N}$ of Similarity Search Across Datasets for Different Latent Dimensions}.}
    \label{fig:search}
\end{figure}

As shown in Fig.~\ref{fig:search}, increasing the number of retrieved samples revealed that non-linear polynomial functions (e.g., KAE ($p=2$, $3$)) achieved competitive results. Especially on more complex datasets like CIFAR10 and CIFAR100, KAEs significantly outperformed FourierKAN and WavKAN, which exhibited lower recall values. This highlights the ability of KAEs to preserve the intrinsic structure of the data in the latent space, making them highly effective for similarity-based retrieval tasks.

\subsubsection{Image Classification}

We further applied the learned latent representations to image classification using a nearest neighbor classifier. For each of the 10,000 test samples, the predicted label was assigned based on the nearest sample in the latent space with the smallest Euclidean distance. We then compared the predicted labels with the ground truth, averaging the classification accuracy across all test samples.

Table~\ref{tab:classification} shows that KAE models, particularly with the non-linear polynomial function ($p=3$), achieved the highest classification accuracy across all datasets. This contrasts with similarity search, where $p=2$ performed best. The difference in performance may due to the nature of each task:
\begin{itemize}
    \item In similarity search, KAE ($p=2$) better preserves the distance relationships between neighbors, which is crucial for retrieving similar samples.

    \item In image classification, KAE ($p=3$) captures more complex, non-linear relationships, enabling better discrimination between class boundaries and improving classification accuracy.
\end{itemize}

\begin{table}[H]
\centering
\caption{\textbf{Classification Accuracy Comparison Across Datasets for Different Latent Dimensions.} The best results are in \textbf{bold} and the second best are \underline{underlined}. The last row shows the improvement of KAE models over standard autoencoders (AE).}
\label{tab:classification}
\setlength{\tabcolsep}{1.7pt}
\small
\begin{spacing}{1.1}
\begin{tabular}{lcccccccc}
\toprule
Dataset & \multicolumn{2}{c}{MNIST} & \multicolumn{2}{c}{FashionMNIST} & \multicolumn{2}{c}{CIFAR10} & \multicolumn{2}{c}{CIFAR100} \\ 
\cmidrule(rr){2-3} \cmidrule(rr){4-5} \cmidrule(rr){6-7} \cmidrule(rr){8-9}
Dimension & 16 & 32 & 16 & 32 & 16 & 32 & 16 & 32 \\ \hline
AE & 0.853$_{\pm \text{0.014}}$ & 0.916$_{\pm \text{0.007}}$ & 0.737$_{\pm \text{0.007}}$ & 0.781$_{\pm \text{0.004}}$ & 0.243$_{\pm \text{0.010}}$ & 0.283$_{\pm \text{0.006}}$ & 0.076$_{\pm \text{0.005}}$ & 0.101$_{\pm \text{0.003}}$ \\
KAN & 0.883$_{\pm \text{0.012}}$ & 0.931$_{\pm \text{0.006}}$ & 0.752$_{\pm \text{0.007}}$ & 0.786$_{\pm \text{0.002}}$ & 0.244$_{\pm \text{0.006}}$ & 0.290$_{\pm \text{0.004}}$ & 0.080$_{\pm \text{0.003}}$ & 0.110$_{\pm \text{0.003}}$ \\
FourierKAN & 0.859$_{\pm \text{0.047}}$ & 0.947$_{\pm \text{0.012}}$ & 0.735$_{\pm \text{0.008}}$ & 0.795$_{\pm \text{0.007}}$ & 0.164$_{\pm \text{0.016}}$ & 0.246$_{\pm \text{0.017}}$ & 0.040$_{\pm \text{0.007}}$ & 0.085$_{\pm \text{0.009}}$ \\
WavKAN & 0.649$_{\pm \text{0.056}}$ & 0.887$_{\pm \text{0.012}}$ & 0.679$_{\pm \text{0.009}}$ & 0.751$_{\pm \text{0.004}}$ & 0.189$_{\pm \text{0.008}}$ & 0.272$_{\pm \text{0.005}}$ & 0.043$_{\pm \text{0.003}}$ & 0.096$_{\pm \text{0.003}}$ \\
\rowcolor{lightblue}
KAE (p=1) & 0.802$_{\pm \text{0.013}}$ & 0.868$_{\pm \text{0.008}}$ & 0.751$_{\pm \text{0.005}}$ & 0.773$_{\pm \text{0.005}}$ & 0.262$_{\pm \text{0.006}}$ & 0.282$_{\pm \text{0.003}}$ & 0.087$_{\pm \text{0.003}}$ & 0.104$_{\pm \text{0.002}}$ \\
\rowcolor{lightblue}
KAE (p=2) & \underline{0.929}$_{\pm \text{0.011}}$ & \underline{0.963}$_{\pm \text{0.002}}$ & \underline{0.801}$_{\pm \text{0.003}}$ & \underline{0.824}$_{\pm \text{0.003}}$ & \underline{0.304}$_{\pm \text{0.010}}$ & \underline{0.338}$_{\pm \text{0.004}}$ & \underline{0.124}$_{\pm \text{0.006}}$ & \underline{0.148}$_{\pm \text{0.002}}$ \\
\rowcolor{lightblue}
KAE (p=3) & \textbf{0.940}$_{\pm \text{0.005}}$ & \textbf{0.964}$_{\pm \text{0.002}}$ & \textbf{0.805}$_{\pm \text{0.004}}$ & \textbf{0.826}$_{\pm \text{0.002}}$ & \textbf{0.315}$_{\pm \text{0.009}}$ & \textbf{0.354}$_{\pm \text{0.004}}$ & \textbf{0.131}$_{\pm \text{0.005}}$ & \textbf{0.154}$_{\pm \text{0.003}}$ \\
Improve & \multicolumn{1}{l}{\color{deepgreen}0.087 $\uparrow$} & \multicolumn{1}{l}{\color{deepgreen}0.048 $\uparrow$} & \multicolumn{1}{l}{\color{deepgreen}0.068 $\uparrow$} & \multicolumn{1}{l}{\color{deepgreen}0.045 $\uparrow$} & \multicolumn{1}{l}{\color{deepgreen}0.072 $\uparrow$} & \multicolumn{1}{l}{\color{deepgreen}0.071 $\uparrow$} & \multicolumn{1}{l}{\color{deepgreen}0.055 $\uparrow$} & \multicolumn{1}{l}{\color{deepgreen}0.053 $\uparrow$} \\
\bottomrule
\end{tabular}
\end{spacing}
\end{table}


\subsubsection{Image Denoising}

Image denoising is a natural extension of autoencoder applications, used to evaluate the robustness of learned models. By adding noise to the input images, we can assess the model's ability to remove noise through compression and reconstruction, using the mean squared error (MSE) between the denoised and original clean images as denoising error. We applied two common types of noise:
\begin{itemize}
    \item \textbf{Gaussian noise}: Added Gaussian noise $\mathcal{N}(0, 0.1^2)$ to the clean images.
    
    \item \textbf{Salt-and-Pepper noise}: Randomly set pixels to 0 or 1 with a probability of 0.05.
\end{itemize}

Results in Table~\ref{tab:denoise} show that both $p=2$ and $p=3$ of KAE consistently achieved the lowest denoising errors across all datasets, mirroring their strong performance in reconstruction. This indicates that the KAE models not only excel in image reconstruction but are also highly effective in removing noise, demonstrating their robustness in preserving the underlying structure of data. The consistent performance across different types of noise further highlights the KAE models' ability to generalize well to varying noise conditions, making them superior to both the standard AE and KAN models.
\vspace{-1mm}

\begin{table}[H]
\centering
\caption{\textbf{Denoising Error Comparison Across Datasets for Different Latent Dimensions.} The best results are in \textbf{bold} and the second best are \underline{underlined}. The last row shows the improvement of KAE models over standard autoencoders (AE).}
\label{tab:denoise}
\setlength{\tabcolsep}{1.7pt}
\small
\begin{spacing}{1.05}
\begin{tabular}{lcccccccc}
\toprule
Dataset & \multicolumn{2}{c}{MNIST} & \multicolumn{2}{c}{FashionMNIST} & \multicolumn{2}{c}{CIFAR10} & \multicolumn{2}{c}{CIFAR100} \\ 
\cmidrule(rr){2-3} \cmidrule(rr){4-5} \cmidrule(rr){6-7} \cmidrule(rr){8-9}
Dimension & 16 & 32 & 16 & 32 & 16 & 32 & 16 & 32 \\ \hline
\multicolumn{9}{c}{\textit{I. Gaussian Noise}} \\
AE & 0.065$_{\pm \text{0.002}}$ & 0.053$_{\pm \text{0.001}}$ & 0.056$_
{\pm \text{0.002}}$ & 0.044$_{\pm \text{0.001}}$ & 0.044$_{\pm \text{0.001}}$ & 0.038$_{\pm \text{0.001}}$ & 0.047$_{\pm \text{0.001}}$ & 0.040$_{\pm \text{0.001}}$ \\
KAN & 0.058$_{\pm \text{0.002}}$ & 0.046$_{\pm \text{0.001}}$ & 0.043$_{\pm \text{0.003}}$ & 0.034$_{\pm \text{0.000}}$ & 0.035$_{\pm \text{0.001}}$ & 0.029$_{\pm \text{0.000}}$ & 0.034$_{\pm \text{0.001}}$ & 0.028$_{\pm \text{0.001}}$ \\
FourierKAN & 0.063$_{\pm \text{0.002}}$ & 0.054$_{\pm \text{0.001}}$ & 0.049$_{\pm \text{0.001}}$ & 0.041$_{\pm \text{0.001}}$ & 0.048$_{\pm \text{0.002}}$ & 0.041$_{\pm \text{0.001}}$ & 0.047$_{\pm \text{0.001}}$ & 0.040$_{\pm \text{0.001}}$ \\
WavKAN & 0.188$_{\pm \text{0.003}}$ & 0.174$_{\pm \text{0.004}}$ & 0.115$_{\pm \text{0.002}}$ & 0.105$_{\pm \text{0.001}}$ & 0.045$_{\pm \text{0.001}}$ & 0.035$_{\pm \text{0.000}}$ & 0.046$_{\pm \text{0.001}}$ & 0.037$_{\pm \text{0.000}}$ \\
\rowcolor{lightblue}
KAE (p=1) & 0.058$_{\pm \text{0.004}}$ & 0.051$_{\pm \text{0.000}}$ & 0.038$_{\pm \text{0.001}}$ & 0.035$_{\pm \text{0.001}}$ & 0.031$_{\pm \text{0.000}}$ & 0.030$_{\pm \text{0.000}}$ & 0.031$_{\pm \text{0.001}}$ & 0.029$_{\pm \text{0.000}}$ \\
\rowcolor{lightblue}
KAE (p=2) & \underline{0.038}$_{\pm \text{0.002}}$ & \underline{0.027}$_{\pm \text{0.001}}$ & \underline{0.030}$_{\pm \text{0.000}}$ & \underline{0.026}$_{\pm \text{0.001}}$ & \textbf{0.027}$_{\pm \text{0.001}}$ & \underline{0.023}$_{\pm \text{0.000}}$ & \textbf{0.026}$_{\pm \text{0.001}}$ & \textbf{0.022}$_{\pm \text{0.000}}$ \\
\rowcolor{lightblue}
KAE (p=3) & \textbf{0.034}$_{\pm \text{0.001}}$ & \textbf{0.025}$_{\pm \text{0.001}}$ & \textbf{0.029}$_{\pm \text{0.001}}$ & \textbf{0.025}$_{\pm \text{0.000}}$ & \textbf{0.027}$_{\pm \text{0.001}}$ & \textbf{0.022}$_{\pm \text{0.000}}$ & \textbf{0.026}$_{\pm \text{0.001}}$ & \textbf{0.022}$_{\pm \text{0.000}}$ \\
Improve & \multicolumn{1}{l}{\color{deepgreen}0.031 $\downarrow$} & \multicolumn{1}{l}{\color{deepgreen}0.028 $\downarrow$} & \multicolumn{1}{l}{\color{deepgreen}0.027 $\downarrow$} & \multicolumn{1}{l}{\color{deepgreen}0.019 $\downarrow$} & \multicolumn{1}{l}{\color{deepgreen}0.017 $\downarrow$} & \multicolumn{1}{l}{\color{deepgreen}0.016 $\downarrow$} & \multicolumn{1}{l}{\color{deepgreen}0.021 $\downarrow$} & \multicolumn{1}{l}{\color{deepgreen}0.018 $\downarrow$} \\
\hline

\multicolumn{9}{c}{\textit{II. Salt-and-Pepper Noise}} \\
AE & 0.092$_{\pm \text{0.002}}$ & 0.082$_{\pm \text{0.001}}$ & 0.078$_{\pm \text{0.001}}$ & 0.069$_{\pm \text{0.001}}$ & 0.058$_{\pm \text{0.001}}$ & 0.053$_{\pm \text{0.001}}$ & 0.061$_{\pm \text{0.001}}$ & 0.055$_{\pm \text{0.001}}$ \\
KAN & 0.086$_{\pm \text{0.002}}$ & 0.075$_{\pm \text{0.001}}$ & 0.067$_{\pm \text{0.002}}$ & 0.060$_{\pm \text{0.000}}$ & 0.050$_{\pm \text{0.001}}$ & 0.045$_{\pm \text{0.000}}$ & 0.050$_{\pm \text{0.001}}$ & 0.045$_{\pm \text{0.000}}$ \\
FourierKAN & 0.087$_{\pm \text{0.001}}$ & 0.080$_{\pm \text{0.002}}$ & 0.071$_{\pm \text{0.001}}$ & 0.065$_{\pm \text{0.001}}$ & 0.065$_{\pm \text{0.001}}$ & 0.059$_{\pm \text{0.001}}$ & 0.064$_{\pm \text{0.001}}$ & 0.059$_{\pm \text{0.001}}$ \\
WavKAN & 0.207$_{\pm \text{0.006}}$ & 0.199$_{\pm \text{0.013}}$ & 0.139$_{\pm \text{0.005}}$ & 0.127$_{\pm \text{0.003}}$ & 0.059$_{\pm \text{0.001}}$ & 0.051$_{\pm \text{0.000}}$ & 0.061$_{\pm \text{0.001}}$ & 0.052$_{\pm \text{0.000}}$ \\
\rowcolor{lightblue}
KAE (p=1) & 0.085$_{\pm \text{0.003}}$ & 0.080$_{\pm \text{0.000}}$ & 0.063$_{\pm \text{0.001}}$ & 0.061$_{\pm \text{0.001}}$ & 0.047$_{\pm \text{0.000}}$ & 0.046$_{\pm \text{0.000}}$ & 0.048$_{\pm \text{0.000}}$ & 0.046$_{\pm \text{0.000}}$ \\
\rowcolor{lightblue}
KAE (p=2) & \underline{0.070}$_{\pm \text{0.002}}$ & \textbf{0.061}$_{\pm \text{0.001}}$ & \textbf{0.057}$_{\pm \text{0.000}}$ & \underline{0.054}$_{\pm \text{0.000}}$ & \textbf{0.044}$_{\pm \text{0.001}}$ & \textbf{0.040}$_{\pm \text{0.000}}$ & \textbf{0.044}$_{\pm \text{0.000}}$ & \textbf{0.040}$_{\pm \text{0.000}}$ \\
\rowcolor{lightblue}
KAE (p=3) & \textbf{0.068}$_{\pm \text{0.001}}$ & \textbf{0.061}$_{\pm \text{0.001}}$ & \textbf{0.057}$_{\pm \text{0.001}}$ & \textbf{0.053}$_{\pm \text{0.000}}$ & \textbf{0.044}$_{\pm \text{0.001}}$ & \textbf{0.040}$_{\pm \text{0.000}}$ & \textbf{0.044}$_{\pm \text{0.000}}$ & \textbf{0.040}$_{\pm \text{0.000}}$ \\
Improve & \multicolumn{1}{l}{\color{deepgreen}0.024 $\downarrow$} & \multicolumn{1}{l}{\color{deepgreen}0.021 $\downarrow$} & \multicolumn{1}{l}{\color{deepgreen}0.021 $\downarrow$} & \multicolumn{1}{l}{\color{deepgreen}0.016 $\downarrow$} & \multicolumn{1}{l}{\color{deepgreen}0.014 $\downarrow$} & \multicolumn{1}{l}{\color{deepgreen}0.013 $\downarrow$} & \multicolumn{1}{l}{\color{deepgreen}0.017 $\downarrow$} & \multicolumn{1}{l}{\color{deepgreen}0.015 $\downarrow$} \\
\bottomrule
\end{tabular}
\end{spacing}
\end{table}


\subsection{Performance Analysis}
\label{sec:evaluation:analysis}

\begin{figure}[H]
    \centering
    \includegraphics[width=0.98\linewidth]{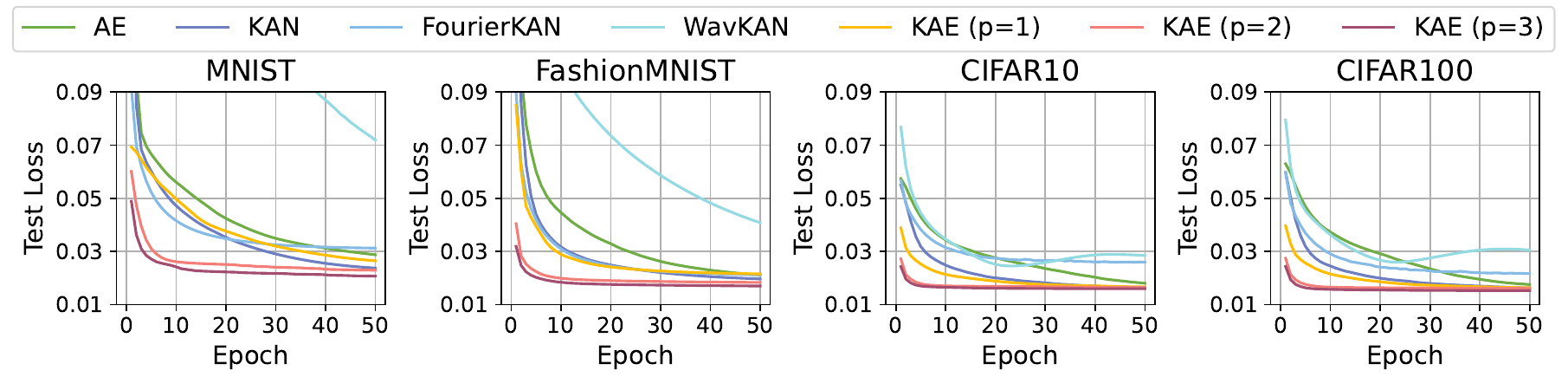}
    \vspace{-2mm}
    
    \caption{\textbf{Convergence Analysis of Test Loss Across Datasets for Latent Dimension $\bm{d_{\text{latent}}=16}$}.}
    \label{fig:loss}
\end{figure}

\textbf{Convergence Analysis.} We measured the test loss as the average MSE between the reconstructed and original data in the test set. Fig.~\ref{fig:loss} illustrates the faster convergence of our KAE models with $p=2$ or $3$, which converge within approximately 10 epochs and achieve the lowest test loss. In contrast, other models, particularly WavKAN and AE, struggle to converge even after 50 epochs.

\begin{wrapfigure}{r}{0.55\textwidth}
    \centering
    \includegraphics[width=0.55\textwidth]{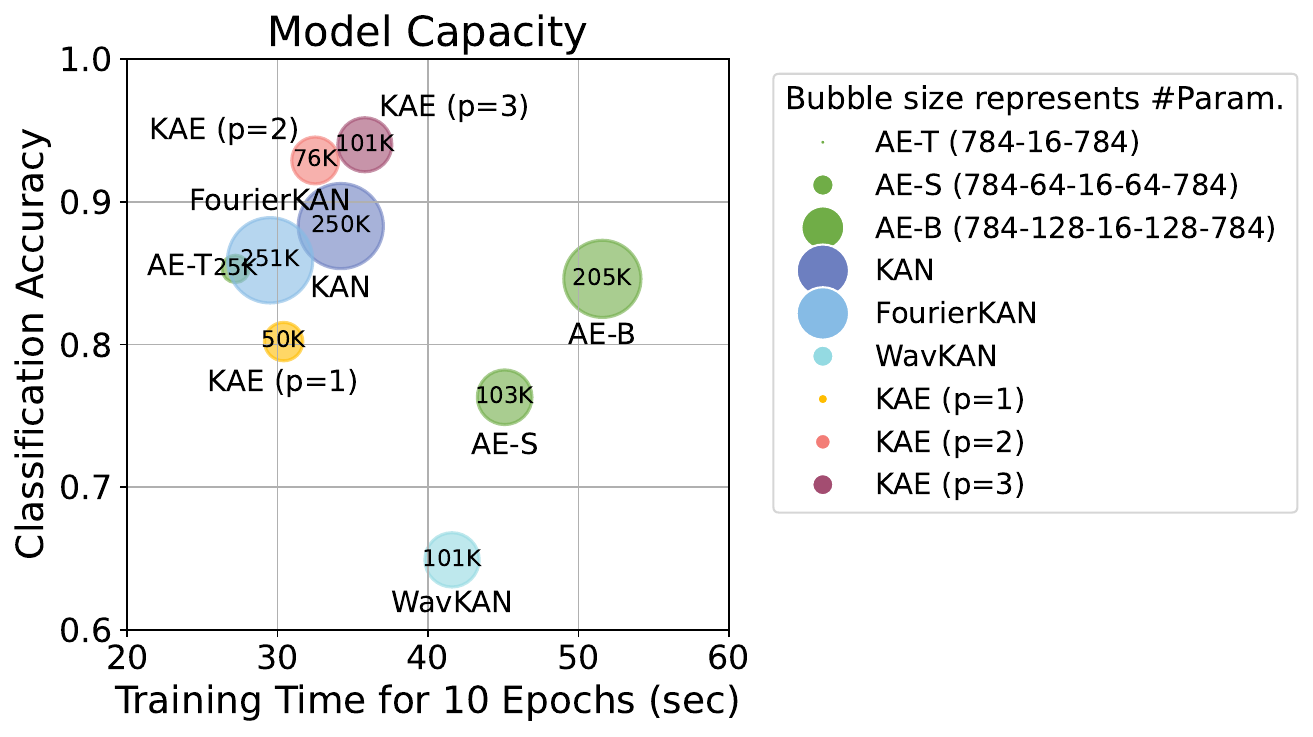}
    \vspace{-5mm}
    
    \caption{\textbf{Model Capacity Analysis on the MNIST Dataset with Latent Dimension $\bm{d_{\text{latent}}=16}$.} Bubble size represents the number of learnable parameters. For AE models, T = Tiny, S = Small, and B = Base.}
    \label{fig:capacity}
    \vspace{-4mm}
\end{wrapfigure}
\textbf{Model Capacity Analysis.} As shown in Fig.~\ref{fig:capacity}, our KAE models strike a balance between efficiency and accuracy, with the following key observations:
\begin{itemize} 
    \item \textbf{Training Efficiency}: While KAE models are not the fastest, they complete training in 30-36 seconds, only marginally slower than the fastest models, and significantly faster than two models with similar parameter counts, i.e., WavKAN and AE-S.

    \item \textbf{Classification Accuracy}: KAE models with \(p=2,3\) use much fewer parameters (75-101K) while achieving higher accuracy compared to KAN (250K) and FourierKAN (251K).

    \item \textbf{Model Parameters}: When comparing models with similar parameter counts, KAE outperforms both WavKAN and AE-S in terms of training speed and performance. Even when the parameters of AE models are doubled, KAE still surpasses AE-B.
\end{itemize}


\section{Conclusion}
\label{sec:conclusion}

In this paper, we introduced the Kolmogorov-Arnold Auto-Encoder (KAE), which integrates the Kolmogorov-Arnold Network (KAN) with autoencoders (AEs) to create a useful and flexible framework for representation learning. By incorporating KAN’s learnable polynomial activation functions into the AE structure, KAE effectively captures complex, non-linear relationships in the data, outperforming standard AEs. Our experiments on benchmark datasets highlight KAE’s superiority in reconstruction quality and downstream applications such as similarity search, image classification, and image denoising. Our analysis further demonstrates KAE’s greater capacity, especially through its learned polynomial activation functions.

Looking ahead, future work will focus on several key directions. First, we plan to combine the KAE architecture with autoencoder variants, such as Variational Auto-Encoders (VAE) \citep{kingma2013vae}. Second, we aim to scale KAE to deeper architectures by incorporating KAN-based CNN layers and techniques like residual connections, enabling it to handle more complex datasets. Third, exploring alternative activation functions enhances the model's flexibility. Finally, applying KAE to challenging real-world tasks, such as image generation, will offer deeper insights into its robustness and adaptability, further demonstrating its potential in practical applications.


\newpage
\bibliography{main}
\bibliographystyle{iclr2025_conference}

\end{document}